\documentclass{article}

\usepackage{neurips_data_2022}
\usepackage{amsmath}
\usepackage{amssymb}
\usepackage{booktabs}
\usepackage{graphicx}
\usepackage{multirow}
\usepackage{times}
\usepackage{soul}
\usepackage{url}
\usepackage[utf8]{inputenc}
\usepackage{enumitem}
\usepackage{makecell}
\usepackage{pifont}
\usepackage{CJKutf8}
\usepackage{caption}
\usepackage{subcaption}

\newcommand{\ch}[1]{\begin{CJK}{UTF8}{gbsn}{#1}\end{CJK}}

\usepackage[dvipsnames, svgnames, x11names]{xcolor}

\usepackage[utf8]{inputenc} 
\usepackage[T1]{fontenc}    
\usepackage{hyperref}       
\usepackage{amsfonts}       
\usepackage{nicefrac}       
\usepackage{microtype}      
\usepackage{xcolor}         

\title{Benchmarking Chinese Text Recognition: \\
Datasets, Baselines, and an Empirical Study}

\author{%
  Haiyang Yu$^{*} $, Jingye Chen\thanks{Equal contribution \newline \hspace*{0.40cm}  \dag Corresponding author}, Bin Li$^{\dag}$, Jianqi Ma, Mengnan Guan, Xixi Xu,
  \\
  \textbf{Xiaocong Wang, Shaobo Qu,  Xiangyang Xue}$^{\dag}$ \\
  Shanghai Key Laboratory of Intelligent Information Processing\\
  School of Computer Science, Fudan University\\
}

\begin{document}

\maketitle

\begin{abstract}
The flourishing blossom of deep learning has witnessed the rapid development of text recognition in recent years. However, the existing text recognition methods are mainly proposed for English texts. As another widely-spoken language, Chinese text recognition (CTR) in all ways has extensive application markets. Based on our observations, we attribute the scarce attention on CTR to the lack of reasonable dataset construction standards, unified evaluation protocols, and results of the existing baselines. To fill this gap, we manually collect CTR datasets from publicly available competitions, projects, and papers. According to application scenarios, we divide the collected datasets into four categories including scene, web, document, and handwriting datasets. Besides, we standardize the evaluation protocols in CTR. With unified evaluation protocols, we evaluate a series of representative text recognition methods on the collected datasets to provide baselines. The experimental results indicate that the performance of baselines on CTR datasets is not as good as that on English datasets due to the characteristics of Chinese texts that are quite different from the Latin alphabet. Moreover, we observe that by introducing radical-level supervision as an auxiliary task, the performance of baselines can be further boosted. The code and datasets are made publicly available at \url{https://github.com/FudanVI/benchmarking-chinese-text-recognition}.

\end{abstract}

\section{Introduction}
In recent years, text recognition has attracted extensive attention due to its wide applications such as autonomous driving \cite{zhang2021character,qian2015robust}, document retrieval \cite{yang2019simple,maekawa2019improving}, signature identification \cite{ren2020st,poddar2020offline}, etc. However, the existing text recognition methods mainly focus on English texts \cite{shi2016end,shi2018aster,cheng2017focusing,cheng2018aon,bai2018edit,luo2019moran,li2019show,qiao2020seed,he2016reading,wan2020textscanner,sheng2019nrtr,wan2020vocabulary} while ignoring the huge market of Chinese text recognition (CTR). Specifically, Chinese is the most spoken language across the world with 1.31 billion speakers, meaning that CTR and its downstream tasks will certainly have a crucial impact on this population.


Based on our observations, we summarize three potential reasons for the scarce attention to Chinese text recognition: 1) \textbf{Lack of reasonable dataset construction standards.} Ideally, based on the given quadrangle boxes, we can crop the text regions along the annotation points and then rectify it to a horizontal-oriented image, which can effectively eliminate useless background areas compared to those methods directly using the minimum circumscribed horizontal boxes. Therefore, different cropping methods will lead to unfair comparison. Also, datasets collected from different environments show much variance in appearance. Finding a reasonable splitting strategy is also beneficial for more effective research. Hence, the standard of dataset construction should be taken into consideration. 2) \textbf{Lack of unified evaluation protocols.} Generally, when evaluating recognition models in English text datasets, researchers usually convert the uppercase to lowercase by default. However, there do not exist any unified evaluation protocols for CTR. For example, researchers may feel confused that whether full-width and half-width characters, simplified and traditional characters should be treated as the same character. Additionally, the evaluation metrics (\textit{e.g.}, normalized edit distance and accuracy rate) are inconsistent across CTR papers. Therefore, unified evaluation protocols are urgently needed for fairly assessing CTR methods. 3) \textbf{Lack of experimental results of the existing baselines.} The existing text recognition methods are mainly evaluated on English text datasets like IIIT5K \cite{mishra2012top}, IC03 \cite{lucas2005icdar}, IC13 \cite{karatzas2013icdar}, etc. Although few methods attempt to experiment on Chinese datasets, the details of dataset construction are not clearly explained in the corresponding papers, which makes it difficult for other researchers to use as CTR baselines.

\begin{figure*}[t]
    \centering
    \includegraphics[width=0.90\textwidth]{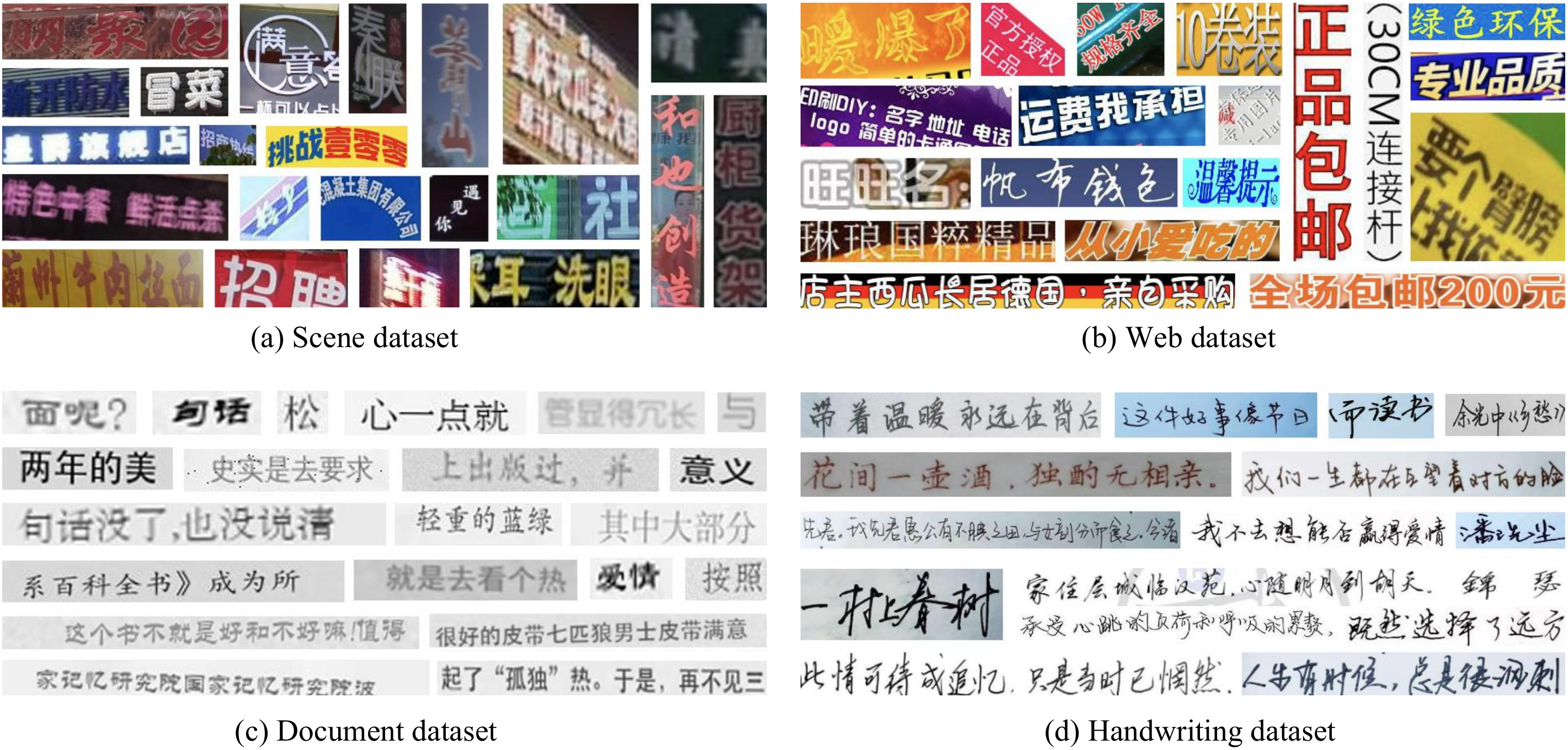}
    \caption{Some samples in the scene, web, document, and handwriting datasets.}
    \label{fig:dataset introduction}
\end{figure*}

In this paper, we seek to construct a benchmark for CTR to fill this gap. We first derive the existing CTR datasets from the public competitions, papers, and projects, resulting in four categories, \textit{i.e.}, scene, web, document, and handwriting (Some examples of each dataset are shown in Figure \ref{fig:dataset introduction}). We also demonstrate the rationality of the dataset construction in terms of each category's recognizability which is calibrated by humans. Then we manually divide each dataset into training, validation, and testing sets at a reasonable ratio. The validation set aims to fairly compare the existing methods, \textit{i.e.}, making sure that the best hyper-parameters is chosen based on the validation set so as to avoid testing-set-oriented tuning. Besides, we reproduce the results of some representative text recognition methods on the collected dataset as baselines. The experimental results indicate that some SOTA methods originally proposed for English texts do not perform as well on CTR datasets as on English datasets. Through analysis, a possible reason is that some characteristics of Chinese texts put obstacles to the existing methods. In view of the complex inner structures of Chinese characters, we introduce the radical-level supervision in a multi-task fashion for better recognition. Overall, our contributions can be listed as follows:

\begin{itemize}
    \item We manually collect CTR datasets from public competitions, papers, and projects. Then we divide them into four categories, \textit{i.e.}, scene, web, document, and handwriting datasets. We further split each dataset into training, evaluation, and testing sets at a reasonable ratio.
    \item We standardize the evaluation protocols to fairly compare existing text recognition methods.
    \item Based on the collected datasets and standardized evaluation protocols, we reproduce the results of a series of baselines, then analyze the performance of the baselines in detail.
    \item We introduce the radical-level supervision in CTR to generally improve the performance of existing attention-based recognizers.
\end{itemize}

\begin{figure*}[t]
    \centering
    \includegraphics[width=1.00\textwidth]{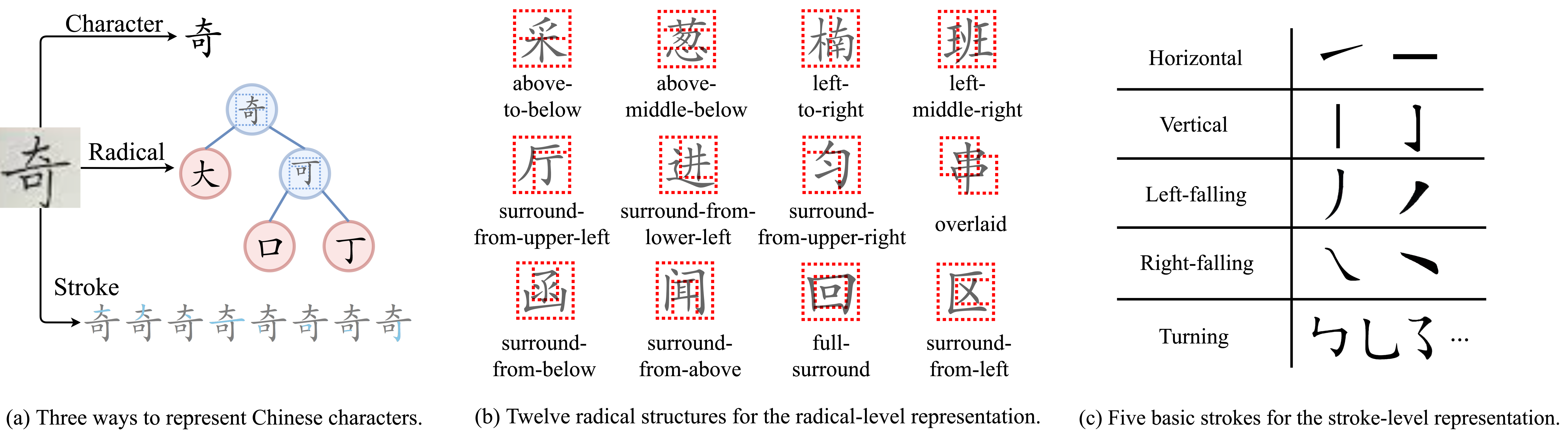}
    \caption{The preliminary knowledge of representing Chinese characters.}
    \label{fig:structure_and_stroke}
\end{figure*}

\section{Preliminaries}

\subsection{Hierarchical Representations for Chinese Characters}
\label{2.1}
Here we introduce three representations of Chinese characters (see the example ``\ch{奇}'' in Figure \ref{fig:structure_and_stroke}(a)), \textit{i.e.}, in character level, radical level, and stroke level.

\textbf{Character level.} According to the Chinese national standard GB18030-2005\footnote{https://zh.wikipedia.org/wiki/GB\_18030}, the number of Chinese characters is 70,244 in total, where 3,755 characters are Level-1 commonly-used characters.

\textbf{Radical level.} According to the Unicode standards for idiographic  description characters\footnote{https://unicode.org/charts/PDF/U2FF0.pdf}, there are 12 radical structures (see Figure \ref{fig:structure_and_stroke}(b)) and 514 radicals in the Level-1 commonly-used Chinese characters. For the 3,755 commonly-used Chinese characters, the radical-level representation can effectively reduce the size of alphabet from 3,755 to 526.

\textbf{Stroke level.} According to Unicode Han Database\footnote{http://www.unicode.org}, each Chinese character can be decomposed into a stroke sequence. There are five basic categories of strokes (\textit{e.g.}, horizontal, vertical, left-falling, right-falling, and turning), each of which contains several instances (see Figure \ref{fig:structure_and_stroke}(c)).




\subsection{Characteristics of Chinese Texts}
\label{characteristicsOfchinese}

\begin{figure*}[t]
    \centering
    \includegraphics[width=1.00\textwidth]{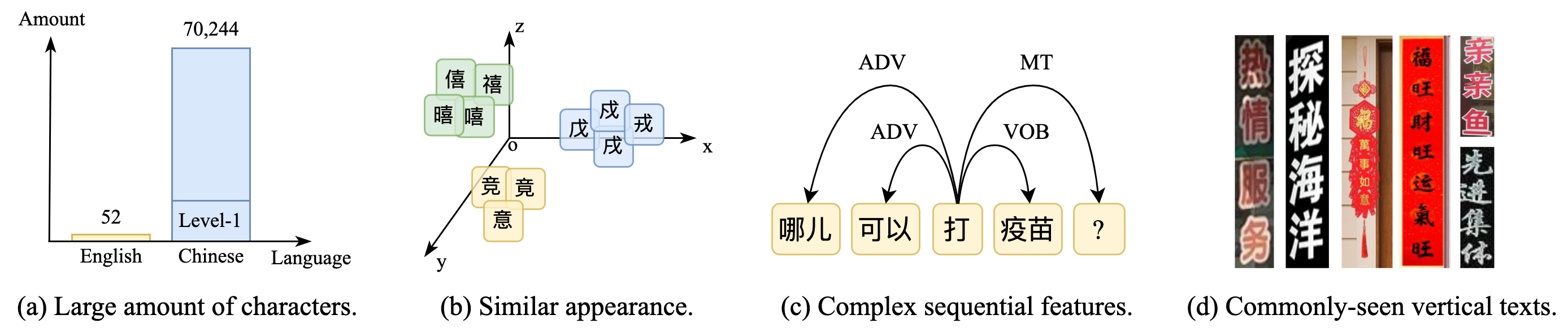}
    \caption{Characteristics of Chinese texts that are distinct from English texts.}
    \label{fig:five properties}
\end{figure*}

It is acknowledged in the community that Chinese texts are harder to recognize compared with English texts \cite{chen2021zero,wang2018denseran}. To explore the intrinsic reasons, here we analyze the characteristics of Chinese texts that are distinct from English texts:

\textbf{Large amount of characters.} According to the national standard GB18030-2005, the number of Chinese characters is 70,244 (including 3,755 commonly-used Level-1 characters). It is much larger than the scale of English characters, which only contains 26 uppercase and 26 lowercase letters (see Figure \ref{fig:five properties}(a)). On one hand, performing large-scale classification is essentially a difficult task \cite{deng2010does}. On the other hand, when experimenting on CTR datasets, recognizers are likely to encounter challenging zero-shot problems, \textit{i.e.}, the characters to be tested may be absent from the training set.

\textbf{Similar appearance.} Compared with English letters, there are considerable Chinese character clusters with similar appearance (see Figure \ref{fig:five properties}(b)). For example, the difference between ``\ch{戌}'' and ``\ch{戍}'' only lies in a tiny stroke. They are hard to distinguish even for human eyes, which indeed burdens the existing text recognition methods.


\textbf{Complicated sequential patterns.} We observe that most samples in the existing English benchmarks are at the word level. Since there is a space between two English words, humans tend to split them apart during the labeling process for detection. Besides, there are inherent statistical patterns within English words (\textit{e.g.}, ``abl’’ is more likely followed by ``e’’ ), thus helping recognizers better capture the sequential patterns. On the contrary, Chinese texts are more likely to appear in phrases or sentences. Under this circumstance, there are complicated dependencies among Chinese characters in terms of the part of speech (see Figure \ref{fig:five properties}(c)), which indeed brings difficulties for recognizers to learn sequential patterns.

\textbf{Commonly-seen vertical texts.} Compared with English texts, Chinese texts are more likely to appear vertically due to the commonly-used traditional couplets or signboards in natural scenes. On the contrary, there are few vertical English texts due to people’s inherent reading habits (see Figure \ref{fig:five properties}(d)).

\textbf{Complex inner structures}. As introduced in \ref{2.1}, different from single-component characters in English, most of Chinese characters are multi-component (see Figure \ref{fig:structure_and_stroke}(a)), \textit{i.e.}, each Chinese character can be decomposed into several radicals organized by radical structures. More complex inner structures makes Chinese character classification more difficult.

\section{Datasets}



\subsection{Details of Datasets}
\textbf{Scene dataset.}
From the competitions, papers, and projects, we have derived a series of scene datasets, including RCTW \cite{shi2017icdar2017}, ReCTS \cite{zhang2019icdar}, LSVT \cite{sun2019icdar}, ArT \cite{chng2019icdar2019}, and CTW \cite{yuan2019large}. The details of each dataset are introduced as follows:

\begin{itemize}
    \item \textbf{RCTW} \cite{shi2017icdar2017}: It provides 12,263 annotated Chinese text images from natural scenes. We derive 44,420 text images from the training set and use them in our benchmark. The testing set of RCTW is not used as the text labels are not available.

    \item \textbf{ReCTS} \cite{zhang2019icdar}: It provides 25,000 annotated street-view Chinese text images, mainly derived from signboards. We only adopt the training set and crop 107,657 text samples in total for our benchmark.

    \item \textbf{LSVT} \cite{sun2019icdar}: It is a large scale Chinese and English scene text dataset, providing 50,000 full-labeled (polygon boxes and text labels) and 400,000 partial-labeled (only one text instance each image) samples. We only utilize the full-labeled training set and crop 243,063 text images for our benchmark.

    \item \textbf{ArT} \cite{chng2019icdar2019}: It contains text samples captured in natural scenes with various text layouts (\textit{e.g.}, rotated text and curved texts). Here we obtain 49,951 cropped text images from the training set, and use them in our benchmark.

    \item \textbf{CTW} \cite{yuan2019large}: It contains annotated 30,000 street view images with rich diversity including raised texts, occluded texts, poorly-illuminated texts, etc. Also, it provides not only character boxes and labels, but also character attributes like background complexity, appearance, etc. Here we crop 191,364 text images from both the training and testing sets.
\end{itemize}

We combine all the above-mentioned datasets, resulting in 636,455 text samples. We randomly shuffle these samples and split them at a ratio of 8:1:1, leading to 509,164 samples for training, 63,645 samples for validation, and 63,646 samples for testing.

\textbf{Web dataset.} To collect the web dataset, we utilize MTWI \cite{he2018icpr2018} that contains 20,000 Chinese and English web text images from 17 different categories on the \textit{Taobao} website. The text samples are appeared in various scenes, typography and designs. We derive 140,589 text images from the training set, and manually divide them at a ratio of 8:1:1, resulting in 112,471 samples for training, 14,059 samples for validation, and 14,059 samples for testing.

\textbf{Document dataset.} We use the public repository Text Render\footnote{https://github.com/Sanster/text\_renderer} to generate some document-style synthetic text images. More specifically, we uniformly sample the length of text varying from 1 to 15. The corpus comes from \textit{Wiki}, \textit{Films}, \textit{Amazon}, and \textit{Baike}. The dataset contains 500,000 in total and is randomly divided into training, validation, and testing sets with a proportion of 8:1:1 (400,000 \textit{v.s.} 50,000 \textit{v.s.} 50,000).

\textbf{Handwriting dataset.} We collect the handwriting dataset based on SCUT-HCCDoc \cite{zhang2020scut}, which captures the Chinese handwritten images with cameras in unconstrained environments. Following the official settings, we derive 93,254 text images for training and 23,389 for testing. To pursue more rigorous research, we manually split the original training set into two sets at a ratio of 4:1, resulting in 74,603 samples for training and 18,651 samples for validation. For convenience, we continue to use the original testing set.

\textbf{Licenses of adopted datasets.} For the datasets RCTW \cite{shi2017icdar2017}, ReCTS \cite{zhang2019icdar}, ArT \cite{chng2019icdar2019}, LSVT \cite{sun2019icdar}, and MTWI \cite{he2018icpr2018}, there were no relevant open-source licenses when these datasets were published. Therefore, we have contacted the owners of these datasets, and confirmed that they can be used for academic research. The dataset CTW \cite{yuan2019large} uses the license Attribution-NonCommercial-ShareAlike 4.0 International, termed CC BY-NC-SA 4.0 \footnote{https://creativecommons.org/licenses/by-nc-sa/4.0/}, where users are authorized to copy and redistribute the material in any medium or format. In the GitHub repository of SCUT-HCCDoc \cite{zhang2020scut} \footnote{https://github.com/HCIILAB/SCUT-HCCDoc\_Dataset\_Release}, the authors have stated that this dataset can only be used for non-commercial research purpose. Also, we have contacted the dataset's authors and obtained permission for academic research.

\begin{figure}[t]
\centering
\begin{minipage}[t]{0.49\textwidth}
\centering
\includegraphics[width=7.0cm]{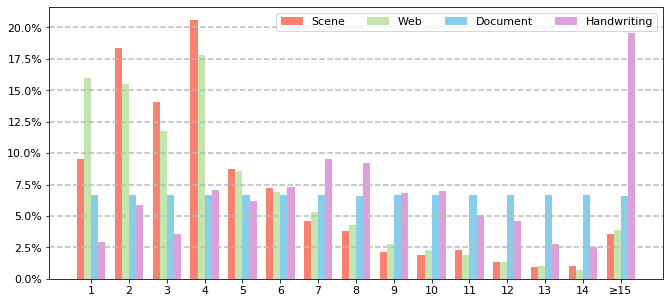}
\caption{The length distribution.}\label{fig:length}
\end{minipage}
\begin{minipage}[t]{0.49\textwidth}
\centering
\includegraphics[width=7.0cm]{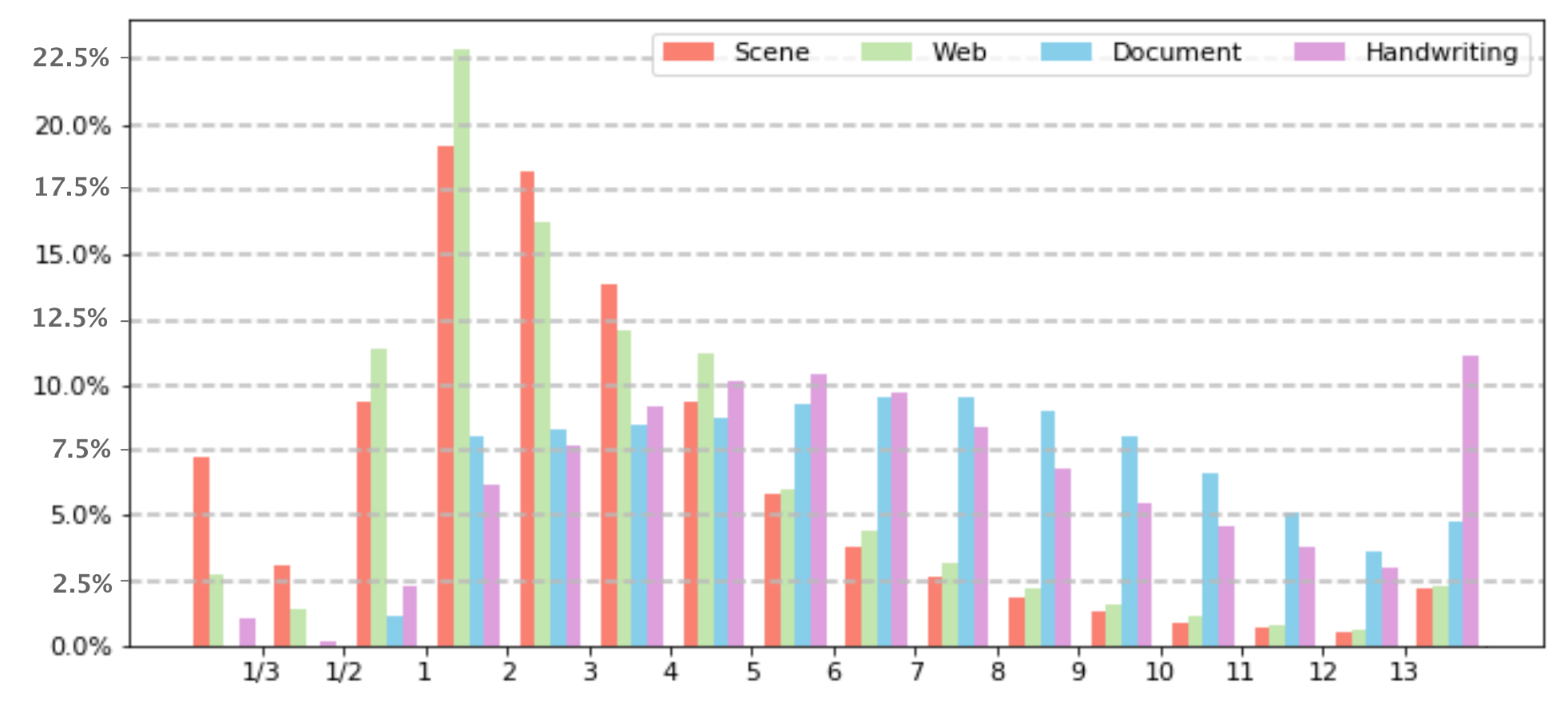}
\caption{The aspect ratio distribution.}\label{fig:ratio}
\end{minipage}
\end{figure}

\begin{table*}[t]
\small
\caption{The statistical results of the alphabet size and the amount of characters. ``Chinese'', ``All'', ``Proportion'' denote the number of Chinese characters, all characters, the proportion of Chinese characters, respectively. ``ZS'' represents the number of zero-shot characters in the testing sets.
}
\centering
\scalebox{0.98}{\begin{tabular}{c p{1.2cm}<{\centering} p{1.1cm}<{\centering} p{1.1cm}<{\centering} p{1.2cm}<{\centering}
p{0.5cm}<{\centering}
p{2.2cm}<{\centering} p{1.6cm}<{\centering}}
\toprule
 \multirowcell{2}{Dataset} &  \multirowcell{2}{Train} & \multirowcell{2}{Valid} & \multirowcell{2}{Test} &
 \multirowcell{2}{Alphabet \\ Size} & \multirowcell{2}{ZS}&
 \multicolumn{2}{c}{Amount of Characters}\\
 \cmidrule(r){7-8}
~ & ~ & ~ & ~ & ~ & ~ & \makecell[c]{Chinese/All} & \makecell[c]{Proportion} \\
\midrule
Scene & 509,164 & 63,645 & 63,646  & 5,880 & 109 & 2,188K / 3,207K & 68.2\% \\
Web & 112,471 & 14,059 & 14,059  & 4,402 & \; 94 & \ \; 315K / 703K & 44.9\% \\
Document & 400,000 & 50,000 & 50,000 & 4,865 & \; 25 & 3,386K / 3,997K & 84.7\% \\
Handwriting & \; 74,603 & 18,651 & 23,389  & 6,105 & 253  & \ \; 983K / 1,154K & 85.3\% \\

\bottomrule
\end{tabular}
}
\label{tab:alphabet and character}
\end{table*}

\begin{figure*}[t]
    \centering
    \includegraphics[width=1.00\textwidth]{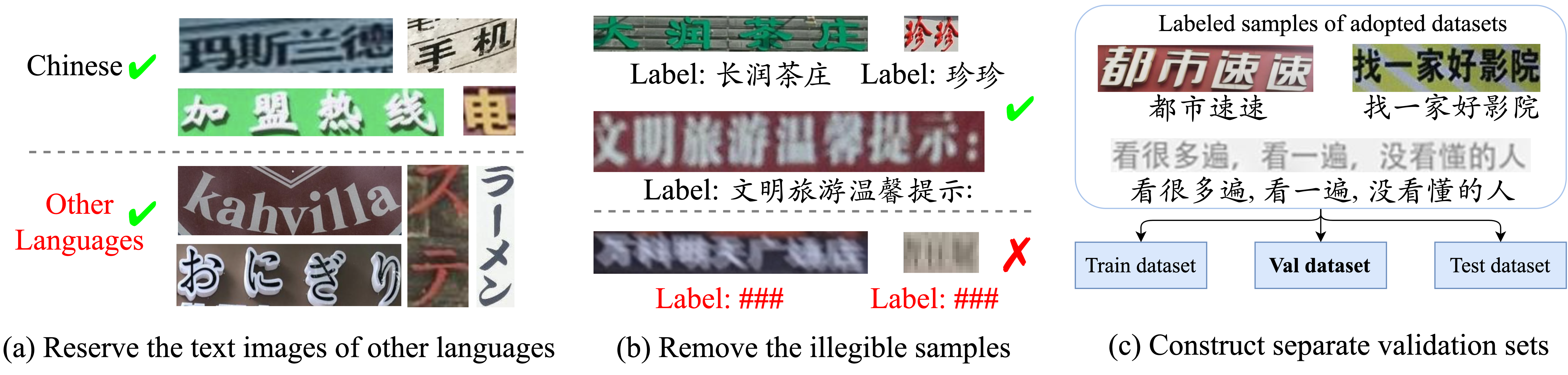}
    \caption{Steps to preprocess the collected datasets.}
    \label{fig:steps}
\end{figure*}

\subsection{Preprocessing}
Here we propose four steps to preprocess the collected four categories of datasets: \textbf{1) Reserve the text images that contain other languages.} We observe that the existing CTR datasets mainly comprise Chinese characters, meanwhile containing a few English characters as well as other languages (see Figure \ref{fig:steps}(a)). Considering the language distributions of text images in natural scenes, we decide to reserve the samples with characters in other languages. \textbf{2) Remove the samples annotated as ``\#\#\#''.} According to the labeling standards of some datasets (\textit{e.g.}, RCTW \cite{shi2017icdar2017} and ReCTS \cite{zhang2019icdar}), the illegible text images are annotated as ``\#\#\#’’ (see Figure \ref{fig:steps}(b)). We observe that there exists heavy blur or occlusion within these samples, which are hard to recognize even for human eyes. Considering that these samples may bring noises to the training process, we decide to remove them from the datasets. \textbf{3) Construct separate validation sets.} Through observations, the existing English text recognition benchmarks usually lack the validation sets. Practically, the testing sets are not available during the training stage and the best model should be chosen according to the performance on the validation sets. Here we take the validation sets into account in pursuit of more rigorous research (see Figure \ref{fig:steps}(c)). \textbf{4) Only collect samples with available text labels.} We observe that the text labels of many testing sets are not publicly available, especially for those competition datasets. Under this circumstance, we only use those samples with available annotated text labels.

\subsection{Analysis of Datasets}

\textbf{Alphabet size and amount of characters.} In Table \ref{tab:alphabet and character}, we can see that the alphabet size and the proportion of Chinese characters vary across the four datasets. For example, most of the web texts are Chinese advertisements with fixed phrase expressions, thus containing fewer characters in the alphabet (4,402 characters). Besides, the web dataset comprises many telephone numbers and English websites, leading to the least proportion of Chinese characters among the four datasets (44.9\%). For the handwriting dataset, most texts are Chinese ancient poems, which contain more rarely-used characters than the other datasets, thus resulting in the largest alphabet (6,105 characters). As a result, the testing set of the handwriting dataset contains the most zero-shot characters (253 zero-shot characters) that are absent from the training set, which further increases the difficulty of recognition.

\textbf{Distributions of text length and aspect ratio}
Figure \ref{fig:length} and Figure \ref{fig:ratio} illustrate the distributions of text length and aspect ratio (\textit{i.e.}, the ratio of width to height). From these figures, we observe that long texts (\textit{e.g.}, length $\ge 10$) appear more frequently in the handwriting dataset, which brings difficulties to the baselines. On the contrary, the text in the scene and web datasets are relatively short, perhaps considering the reading efficiency of passengers, customers, etc. For the ratio distribution, we observe that the scene dataset has more vertical text than others (ratio $\le$ 1) due to the commonly-used couplets and signboards in the wild. By contrast, the handwriting dataset comprises more horizontal texts.

\textbf{Character and word frequency.} As shown in Table \ref{tab:frequency}, we analyze the frequency of characters and words\footnote{We use jieba to split texts into words. The homepage of jieba: https://github.com/fxsjy/jieba} in each dataset and observe some interesting phenomena. For instance, as there are many Chinese brands or state-owned enterprises in natural scenes, the characters \begin{CJK}{UTF8}{gbsn}``中'', ``国'', \end{CJK} and the word \begin{CJK}{UTF8}{gbsn}``中国''\end{CJK} (China) appear frequently in the scene dataset. Additionally, the web dataset contains many high-frequency advertising terms like \begin{CJK}{UTF8}{gbsn}``包邮'' (free shipping), ``正品'' (certified products), etc.\end{CJK} Since most of the samples in the document and handwriting datasets are Chinese sentences, the auxiliary word \begin{CJK}{UTF8}{gbsn}``的''\end{CJK} appears most frequently. In particular, in the handwriting dataset, since some text images are cropped from diaries, there are many characters or words referring to people like \begin{CJK}{UTF8}{gbsn}``我'' (I), ``你'' (you), ``我们'' (we), etc.\end{CJK}


\begin{table*}[t]
\small
\caption{Top six high-frequency characters and words in each dataset (frequency in parentheses).
}
\centering
\scalebox{0.95}{\begin{tabular}{c | r r r r r r r r}
\toprule
 \multirow{2}*{Scene} & \begin{CJK}{UTF8}{gbsn}电\footnotesize{(26,986)}\end{CJK} & \begin{CJK}{UTF8}{gbsn}中\footnotesize{(26,564)}\end{CJK} & \begin{CJK}{UTF8}{gbsn}店\footnotesize{(18,644)}\end{CJK} & \begin{CJK}{UTF8}{gbsn}国\footnotesize{(17,835)}\end{CJK} & \begin{CJK}{UTF8}{gbsn}大\footnotesize{(17,819)}\end{CJK} & \begin{CJK}{UTF8}{gbsn}话\footnotesize{(16,543)}\end{CJK}\\
~ & \begin{CJK}{UTF8}{gbsn}电话\footnotesize{(14,969)}\end{CJK} & \begin{CJK}{UTF8}{gbsn}中国\footnotesize{(5,554)}\end{CJK} & \begin{CJK}{UTF8}{gbsn}有限公司\footnotesize{(5,298)}\end{CJK} & \begin{CJK}{UTF8}{gbsn}杭州\footnotesize{(3,793)}\end{CJK} & \begin{CJK}{UTF8}{gbsn}热线\footnotesize{(3,612)}\end{CJK} & \begin{CJK}{UTF8}{gbsn}中心\footnotesize{(3,505)}\end{CJK} \\
\midrule
 \multirow{2}*{Web} & \begin{CJK}{UTF8}{gbsn}品\footnotesize{(4,077)}\end{CJK} & \begin{CJK}{UTF8}{gbsn}包\footnotesize{(3,104)}\end{CJK} & \begin{CJK}{UTF8}{gbsn}电\footnotesize{(2,682)}\end{CJK} & \begin{CJK}{UTF8}{gbsn}送\footnotesize{(2,597)}\end{CJK} & \begin{CJK}{UTF8}{gbsn}一\footnotesize{(2,493)}\end{CJK} & \begin{CJK}{UTF8}{gbsn}全\footnotesize{(2,413)}\end{CJK}\\
~ & \begin{CJK}{UTF8}{gbsn}包邮\footnotesize{(1,851)}\end{CJK} & \begin{CJK}{UTF8}{gbsn}正品\footnotesize{(1,627)}\end{CJK} & \begin{CJK}{UTF8}{gbsn}电子\footnotesize{(625)}\end{CJK} & \begin{CJK}{UTF8}{gbsn}原装\footnotesize{(602)}\end{CJK} & \begin{CJK}{UTF8}{gbsn}全国\footnotesize{(538)}\end{CJK} & \begin{CJK}{UTF8}{gbsn}品质\footnotesize{(519)}\end{CJK}\\
\midrule
 \multirow{2}*{Document} & \begin{CJK}{UTF8}{gbsn}的\footnotesize{(149,351)}\end{CJK} & \begin{CJK}{UTF8}{gbsn}是\footnotesize{(60,534)}\end{CJK} & \begin{CJK}{UTF8}{gbsn}不\footnotesize{(54,367)}\end{CJK} & \begin{CJK}{UTF8}{gbsn}一\footnotesize{(51,760)}\end{CJK} & \begin{CJK}{UTF8}{gbsn}这\footnotesize{(38,080)}\end{CJK} & \begin{CJK}{UTF8}{gbsn}有\footnotesize{(36,839)}\end{CJK}\\
~ & \begin{CJK}{UTF8}{gbsn}不错\footnotesize{(8,411)}\end{CJK} & \begin{CJK}{UTF8}{gbsn}没有\footnotesize{(7,382)}\end{CJK} & \begin{CJK}{UTF8}{gbsn}故事\footnotesize{(7,263)}\end{CJK} & \begin{CJK}{UTF8}{gbsn}这个\footnotesize{(7,247)}\end{CJK} & \begin{CJK}{UTF8}{gbsn}可以\footnotesize{(5,938)}\end{CJK} & \begin{CJK}{UTF8}{gbsn}一个\footnotesize{(5,263)}\end{CJK}\\
\midrule
 \multirow{2}*{Handwriting} & \begin{CJK}{UTF8}{gbsn}的\footnotesize{(31,845)}\end{CJK} & \begin{CJK}{UTF8}{gbsn}我\footnotesize{(17,543)}\end{CJK} & \begin{CJK}{UTF8}{gbsn}不\footnotesize{(15,663)}\end{CJK} & \begin{CJK}{UTF8}{gbsn}一\footnotesize{(15,028)}\end{CJK} & \begin{CJK}{UTF8}{gbsn}是\footnotesize{(13,257)}\end{CJK} & \begin{CJK}{UTF8}{gbsn}你\footnotesize{(11,773)}\end{CJK} \\
~ & \begin{CJK}{UTF8}{gbsn}自己\footnotesize{(2,474)}\end{CJK} & \begin{CJK}{UTF8}{gbsn}抄写\footnotesize{(1,751)}\end{CJK} & \begin{CJK}{UTF8}{gbsn}我们\footnotesize{(1,739)}\end{CJK} & \begin{CJK}{UTF8}{gbsn}一个\footnotesize{(1,270)}\end{CJK} & \begin{CJK}{UTF8}{gbsn}不是\footnotesize{(1,171)}\end{CJK} & \begin{CJK}{UTF8}{gbsn}如果\footnotesize{(973)}\end{CJK}\\
\bottomrule
\end{tabular}
}
\label{tab:frequency}
\end{table*}

\begin{table*}[t]
\small
\caption{The statistical results of recognizability calibrated by humans (BG for background).
}
\centering
\scalebox{0.97}{\begin{tabular}{c p{1.6cm}<{\centering}   p{2.4cm}<{\centering} p{2.2cm}<{\centering} p{1.6cm}<{\centering} p{1.7cm}<{\centering}}
\toprule
Dataset & Occlusion & Oblique or Curved & BG Confusion & Scribble & Blur \\
\midrule
Scene & \textbf{37(7.4\%)} & 81(16.2\%) & \textbf{93(18.6\%)} & \;\;\; 28(5.6\%) & \textbf{189(37.8\%)} \\
Web & 29(5.8\%) & \textbf{91(18.2\%)} & 67(13.4\%) & \;\;\; 11(2.2\%) & 106(21.2\%) \\
Document & \; 5(1.0\%) & 50(10.0\%) & \;\;\; 1(0.2\%) & \;\;\;\;\; 1(0.2\%) & \;\;\; 42(8.4\%) \\
Handwriting & 11(2.2\%) & \; 24(4.8\%) & \; 30(6.0\%) & \textbf{202(40.4\%)} & \;\;\; 35(7.0\%) \\
\bottomrule
\end{tabular}
}
\label{tab:visual experiment}
\end{table*}

\textbf{Recognizability calibration by humans.} To figure out the factors that impair the recognizability of text images in the four datasets, we invite 20 highly-educated\footnote{Ensure that the participant can recognize all well-written characters.} participants to conduct this experiment. We invite the participants to identify the corresponding reasons (multiple choices) that hamper the recognizability from different perspectives: 1) Occlusion (from foreground), 2) Oblique or Curved (instance-level), 3) Background Confusion (from background), 4) Scribble (character-level), 5) Blur (image source). If none of the reasons is satisfied, we consider the text image recognizable. Overall, we have 500 votes for each dataset (20 participants on 25 image samples per dataset). The statistics are demonstrated in Table \ref{tab:visual experiment}. We can see that, in the scene dataset, the votes of participants mainly focus on ``Occlusion'', ``Background Confusion'', and ``Blur'', indicating that it is the most complicated dataset for humans to recognize. The factors that impair the recognizability mostly come from the wild environment and the way of capturing text images. For the web dataset, the text appearance can be more various through controllable generation, resulting in the most votes for the ``Oblique or Curved'' selection. For the document dataset, the samples are relatively recognizable for humans and thus are less voted. In the handwriting dataset, participants vote the most for ``Scribble'', indicating that the stroke cohesion in the writing is also a basic factor that precludes the recognizability. From the observations above, we can find that the factors that impair the recognizability vary across datasets, and thus motivate us to investigate the performance on each dataset separately.

\section{Baselines}
\label{baseline}
Text recognition has achieved rapid progress over the last decade. According to the major characteristics, text recognition methods can be classified into several categories, including CTC-based methods, rectification-based methods, etc. From these categories, we select as baselines eight \textit{representative} methods, which are mostly used for comparison in the text recognition task.



\textbf{CRNN} (Shi \textit{et al.}, 2016) \cite{shi2016end} is a typical CTC-based method and it is widely used in industry. It sends the text image to a CNN to extract the image features, then adopts a two-layer LSTM to encode the sequential patterns. Finally, the output of LSTM is fed to a CTC~(Connectionist Temperal Classification)~\cite{graves2006connectionist} decoder to maximize the probability of all the paths towards the ground truth.


\textbf{ASTER} (Shi \textit{et al.}, 2018) \cite{shi2018aster} is a typical rectification-based method aiming at tackling irregular text images. It introduces a Spatial Transformer Network (STN)~\cite{jaderberg2015spatial} to rectify the given text image. Then the rectified text image is sent to a CNN and a two-layer LSTM to extract the features. In particular, ASTER takes advantage of the attention mechanism to predict the final text sequence.

\textbf{MORAN} (Luo \textit{et al.}, 2019)  \cite{luo2019moran} is a representative rectification-based method. It first adopts a multi-object rectification network~(MORN) to predict rectified pixel offsets in a weak supervision way (distinct from ASTER that utilizes STN). The output pixel offsets are further used for generating the rectified image, which is further sent to the attention-based decoder~(ASRN) for text recognition.

\textbf{SAR} (Li \textit{et al.}, 2019) \cite{li2019show} is a representative method that takes advantage of 2-D feature maps for more robust decoding. Different from CRNN, ASTER, and MORAN compressing the given image into a 1-D feature map, SAR adopts 2-D attention on the spatial dimension of the feature maps for decoding, resulting in a stronger performance in curved and oblique texts.

\textbf{SEED} (Qiao \textit{et al.}, 2020)  \cite{qiao2020seed} is a representative semantics-based method. It introduces a semantics module to extract global semantic embedding and utilize it to initialize the first hidden state of the decoder. Benefiting from introducing semantics prior to the recognition process, the decoder of SEED shows superiority in recognizing low-quality text images.


\textbf{MASTER} (Lu \textit{et al.}, 2021)
\cite{lu2021master} is an attention-based method that utilizes the self-attention mechanism to learn a more powerful and robust representation for distorted text images. Meanwhile, benefiting from the training parallelization and memory-cache mechanism, this method has a great training efficiency and a high-speed inference.

\textbf{ABINet} (Fang \textit{et al.}, 2021)
\cite{fang2021read} is an autonomous, bidirectional, and iterative method for scene text recognition. The autonomous principle means that the vision model and language model should be learned separately; the bidirectional principle is proposed to capture twice the amount of information; the iterative principle is designed to refine the prediction from visual and linguistic cues. In addition, the authors proposed a self-training method that enables ABINet to learn from unlabeled images.

\textbf{TransOCR} (Chen \textit{et al.}, 2021) \cite{chen2021scene} is one of the representative Transformer-based methods. It is originally designed to provide text priors for the super-resolution task. It employs ResNet-34 \cite{he2016deep} as the encoder and self-attention modules as the decoder. Distinct from the RNN-based decoders, the self-attention modules are more efficient to capture semantic features of the given text images.

Through the experimental results of baselines (the details will be introduced in Section \ref{experiments}), we observe that the performance of baselines on CTR datasets is not as good as that on English datasets due to the characteristics of Chinese texts (see Section \ref{characteristicsOfchinese}). In terms of complex inner structures of Chinese characters, we propose a Pluggable Radical-Aware Branch (PRAB), which can be inserted into any attention-based recognizer, to introduce the radical-level supervision in a multi-task fashion for better recognition. More details about PRAB are shown in Supplementary Materials.

\section{An Empirical Study}
\label{experiments}

\begin{table*}[t]
\small
\caption{The results of the baselines on four datasets. ACC / NED follows the percentage format and decimal format, respectively.
}
\centering
\scalebox{0.95}{\begin{tabular}{p{2.5cm}<{\centering} p{1.7cm}<{\centering} p{1.7cm}<{\centering} p{1.7cm}<{\centering} p{1.7cm}<{\centering} p{1cm}<{\centering} p{1cm}<{\centering}}
\toprule
\multirow{2}*{Method} & \multicolumn{4}{c}{Dataset} & \multirow{2}*{Parameters} & \multirow{2}*{FPS} \\
\cmidrule{2-5}
~  & Scene & Web & Document & Handwriting \\
\midrule
CRNN \cite{shi2016end} & 54.94 / 0.742 & 56.21 / 0.745 & 97.41 / 0.995 & 48.04 / \textbf{0.843} & \textbf{12.4M} & \textbf{751.0} \\

ASTER \cite{shi2018aster} & 59.37 / 0.801 & 57.83 / \textbf{0.782} & 97.59 / 0.995 & 45.90 / 0.819 & 27.2M & 107.3 \\

MORAN \cite{luo2019moran} & 54.68 / 0.710 & 49.64 / 0.679 & 91.66 / 0.984 & 30.24 / 0.651 & 28.5M & 301.5 \\

SAR \cite{li2019show} & 53.80 / 0.738 & 50.49 / 0.705 & 96.23 / 0.993 & 30.95 / 0.732 & 27.8M & 93.1 \\

SEED \cite{qiao2020seed} & 45.37 / 0.708 & 31.35 / 0.571 & 96.08 / 0.992 & 21.10 / 0.555 & 36.1M & 106.6 \\
MASTER \cite{lu2021master} & 62.14 / 0.763 &	53.42 / 0.704 & 82.69 / 0.957 & 18.52 / 0.504 & 62.8M & 16.3 \\
ABINet \cite{fang2021read} & 60.88 / 0.775 & 51.07 / 0.704 & 91.67 / 0.987 & 13.83 / 0.514 & 53.1M & 92.1 \\

TransOCR \cite{chen2021scene} & \textbf{67.81 / 0.817} & \textbf{62.74 / 0.782} & \textbf{97.86 / 0.996} & \textbf{51.67} / 0.835 & 83.9M & 164.6 \\

\bottomrule
\end{tabular}
}
\label{tab:new baseline result}
\end{table*}

\subsection{Experiments}

\textbf{Implementation details}. We adopt the off-the-shelf PyTorch implementations of CRNN \cite{shi2016end}, ASTER \cite{shi2018aster}, MORAN \cite{luo2019moran}, SAR \cite{li2019show}, SEED \cite{qiao2020seed}, MASTER \cite{lu2021master}, ABINet \cite{fang2021read} and TransOCR \cite{chen2021scene} on Github to reproduce the experimental results on the collected CTR datasets. All baseline experiments are conducted on an NVIDIA RTX 2080Ti GPU with 11GB memory. To maintain the time efficiency of existing recognizers and make text images more recognizable for recognizers, we resize the input image into $32 \times 256$ for all experiments. We utilize the validation set of each dataset to choose the best hyper-parameters according to the recognition accuracy, then assess the baselines using the testing set. For convenience, we combine the alphabets of four datasets for all experiments, resulting in an overall alphabet of 7,934 characters. No other strategies like data augmentation and pre-training are used for the baselines.

\textbf{Evaluation protocols.} Practically, the unified evaluation protocols are indispensable for fair comparison. Following the ICDAR2019 ReCTS Competition\footnote{https://rrc.cvc.uab.es/?ch=12\&com=tasks}, we leverage some rules to convert the predictions and labels:  (1) Convert full-width characters to half-width characters; (2) Convert traditional Chinese characters to simplified characters; (3) Convert uppercase letters to lowercase letters; (4) Remove all spaces. After these transformations, we utilize the widely-used metric Accuracy (\textbf{ACC}) to evaluate the baselines. In addition, Normalized Edit Distance (\textbf{NED}) is utilized to comprehensively evaluate the performance of baselines since CTR datasets contain more long text images compared with English text recognition datasets. ACC and NED are both ranged in [0,1]. Higher ACC and NED indicate better performance of the evaluated baseline.

\textbf{Experimental results.} We first analyze the experimental results (see Table \ref{tab:new baseline result}) of different methods. We observe that the performance of CRNN \cite{shi2016end} on each dataset is better than that of those vanilla attention-based recognizers (\textit{i.e.}, MORAN \cite{luo2019moran}, SEED \cite{qiao2020seed}, and SAR \cite{li2019show}), which are vulnerable to the attention drift problem \cite{cheng2017focusing} when encountering text images with longer Chinese texts. Although the performance of CRNN on the scene dataset is not as good as those transformer-based methods, CRNN has fewer parameters and costs less time in inference. Moreover, we notice that SEED does not perform well on all datasets. A possible reason may be that SEED needs to map each text image to its corresponding semantic embedding guided by fastText \cite{bojanowski2017enriching}, whereas Chinese text usually contains complex semantics, thus bringing difficulties to the semantics learning procedure. Obviously, TransOCR \cite{chen2021scene} surpasses all its counterparts as it is capable of modeling the sequential patterns more flexibly. Next, we analyze the experimental results from the perspective of dataset. As shown in Table \ref{tab:new baseline result}, all baselines do not perform well on the handwriting dataset due to the scribble in this dataset. As demonstrated in Table \ref{tab:visual experiment}, almost 40\% of the text samples in the handwriting dataset is marked as ``Scribble''. Practically, the writer may join up or omit some strokes to accelerate the writing speed, which indeed brings difficulties to the existing methods. On the contrary, the recognition accuracy of all baselines can exceed 90.0\% in the document dataset since the text samples in this dataset are more recognizable compared with the other three datasets. Although more samples are contained in the scene dataset, some problems (\textit{e.g.}, occlusion, background confusion, blur, etc.) still challenge the baseline methods, leading to the relatively poor performance on the scene dataset. For the web dataset, the performance of all baselines is lower than that of the scene dataset, which may stem from the scarcity of training samples.

\begin{table*}[t]
\small
\caption{The results of the baselines on hard cases. ACC / NED follows the percentage format and decimal format, respectively.
}
\centering
\scalebox{0.97}{\begin{tabular}{p{2.1cm}<{\centering} p{1.7cm}<{\centering} p{2.4cm}<{\centering} p{2cm}<{\centering} p{1.7cm}<{\centering} p{1.9cm}<{\centering}}
\toprule
Methods & Occlusion & \small{Oblique or Curved} & \small{BG Confusion} & Blur & Vertical Text \\
\midrule
CRNN \cite{shi2016end} & 22.0 / 0.531 & 13.0 / 0.389 & 22.0 / 0.621 & 34.0 / 0.535 & \; 2.0 / 0.164 \\
ASTER \cite{shi2018aster} & 28.0 / \textbf{0.641} & 21.0 / \textbf{0.510} & 32.0 / 0.710 & 39.6 / \textbf{0.658} & 21.0 / \textbf{0.491} \\
MORAN \cite{luo2019moran} & 21.0 / 0.459 & 10.0 / 0.280 & 17.0 / 0.479 & 31.8 / 0.441 & 10.0 / 0.166 \\
SAR \cite{li2019show} & 19.0 / 0.532 & 15.0 / 0.379 & 23.0 / 0.600 & 30.8 / 0.507 & \; 6.0 / 0.223 \\
SEED \cite{qiao2020seed} & 12.0 / 0.469 & \; 5.0 / 0.288 & 22.0 / 0.566 & 28.8 / 0.521 & \; 9.0 / 0.384 \\
MASTER \cite{lu2021master} & 26.0 / 0.509 & 12.0 / 0.314 & 30.0 / 0.587 & 41.2 / 0.532 & 13.0 / 0.199 \\
ABINet \cite{fang2021read} & 20.0 / 0.512 & 14.0 / 0.354 & 19.0 / 0.545 & 38.0 / 0.570 & \; 8.0 / 0.200 \\
TransOCR \cite{chen2021scene} & \textbf{32.0} / 0.581 & \textbf{29.0} / 0.503 & \textbf{44.0} / \textbf{0.741} & \textbf{50.4} / 0.626 & \textbf{32.0} / 0.440 \\
\bottomrule
\end{tabular}
}
\label{tab:hard case}
\end{table*}

\subsection{Discussions}

\textbf{Pretrained on synthetic datasets.} Considering that previous English text recognition methods tend to be trained on large-scale synthetic datasets, we also produce a synthetic CTR dataset to pretrain baseline models for further improving the performance. Details about the synthetic CTR dataset and experimental results are shown in the Supplementary Material. Through extensive experiments, we observe that pretraining baseline models with the synthetic dataset can indeed improve performance in most cases. Although a large synthetic dataset is used for pretraining, baseline models still cannot achieve the expected performance due to the characteristics of Chinese texts.

\textbf{Effectiveness of the PRAB.} Considering the characteristic of Chinese characters that each Chinese character can be decomposed into a specific radical sequence according to cjkvi-ids\footnote{https://github.com/cjkvi/cjkvi-ids}, we propose a Pluggable Radical-Aware Branch (PRAB) to introduce the more fine-grained supervision (\textit{i.e.}, radicals) for each character. We conduct extensive experiments to evaluate the effectiveness of PRAB by equipping it for all attention-based methods mentioned in Section \ref{baseline}. Through the experimental results (shown in the Supplementary Material), we observe that an attention-based recognizer can perform better on four datasets when the proposed PRAB is used to provide radical-level supervision, which validates the effectiveness of PRAB. Specifically, with the help of PRAB, the average accuracy improvement on MORAN, SAR, and TransOCR is 1.83\%, 3.35\%, and 2.33\% respectively.

\textbf{Analysis on hard cases.} We manually select some hard cases (\textit{e.g.}, occlusion, oblique or curved, confusing background, blur, and vertical) and analyze the performance of baselines on these cases\footnote{We manually pick 100 occlusion samples, 100 oblique or curved samples, 100 confusing background samples, 100 vertical samples, 500 blur samples in the scene and web datasets.} Please note that the ``Scribble’’ situation is mainly associated with the handwriting dataset so we do not analyze the ``Scribble’’ case separately. The experimental results on these hard cases are shown in Table \ref{tab:hard case}. In particular, we observe that TransOCR \cite{chen2021scene} can well adapt to each hard case compared with other baselines. Benefiting from the self-attention modules, the recognizer can easily tackle uncommon text layouts like oblique and curved, or mitigate the noise induced by confusing foregrounds (occlusion) or confusing backgrounds. Interestingly, ASTER\cite{shi2018aster} achieves better NED compared with TransOCR \cite{chen2021scene} in most situations as the STN utilized by ASTER increases the recognizability of those hard samples with a certain degree of oblique. Additionally, the CTC-based CRNN \cite{shi2016end} simply transforms original input to 1-D features, thus performing poorly on the vertical text images. Overall, there is still large room for improvement in terms of these hard cases.

\section{Conclusions}
In this paper, we first discuss possible reasons for the scarce attention on Chinese text recognition. To tackle these problems, we collect the publicly available datasets and divide them into scene, web, document, and handwriting datasets, respectively. We also analyze the characteristics of each dataset. Then we standardize the evaluation protocols in CTR (\textit{e.g.}, whether researchers should treat the traditional and simplified Chinese characters as the same character), which enables the researchers without Chinese character knowledge to participate in CTR research. Finally, we adopt eight representative methods as baselines on the collected datasets. Through the empirical study, we observe that combining the Chinese character knowledge is helpful for the Chinese text recognition task, which also provides guidance for future work of CTR.

{\small
\bibliographystyle{unsrt}
\bibliography{neurips_data_2022}
}

\newpage
\appendix

\section{Details of PRAB}
\begin{figure*}[ht]
    \centering
    \includegraphics[width=1.00\textwidth]{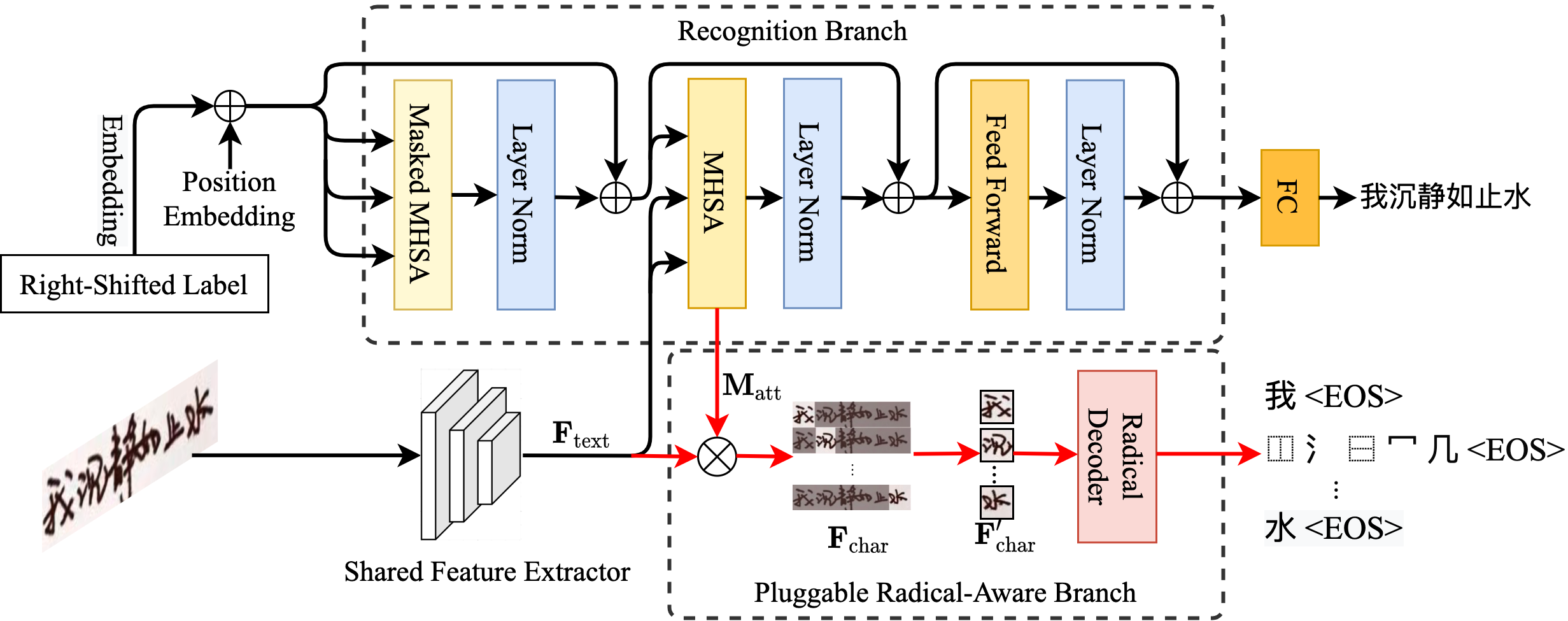}
    \caption{The framework of our method. The data flow indicated by red arrows is only used in the training phase.}
    \label{fig:framework}
\end{figure*}

Through the experimental results of baselines, we observe that the performance of the existing recognizers on collected CTR datasets is not as good as that on English datasets. A possible reason is that the characteristics of Chinese characters are quite different from that of the Latin alphabet. Considering that each Chinese character can be decomposed into a specific radical sequence, we propose a Pluggable Radical-Aware Branch (PRAB) to introduce more fine-grained supervision (\textit{i.e.}, radicals), which enables the recognizer to capture radical-aware features for better recognition. As shown in Figure \ref{fig:framework}, the framework of our method can be divided into three parts: shared feature extractor, recognition branch, and PRAB. The PRAB shares the feature extractor with the recognizer and crops the feature maps of each character with attention masks of the recognizer. Through a transformer-based decoder, each character is decoded into a radical sequence, the supervision of which is off-the-shelf as introduced in \cite{wang2018denseran}.

\subsection{Shared Feature Extractor}
Here, we use TransOCR \cite{chen2021scene} as the baseline, thus ResNet-34 \cite{he2016deep} is employed as the shared feature extractor. To retain more spatial information for the PRAB, we drop the last three down-sampling layers in ResNet-34, thus the dimension of the feature maps $\mathbf{F}_\text{text}$ after the shared feature extractor is $\frac{H}{4} \times \frac{W}{4} \times C$ where $H$ and $W$ denote the height and width of input text images.

\subsection{Recognition Branch}
To providing attention masks for the PRAB, we should adopt a 2-D attention-based recognizer, \textit{i.e.}, TransOCR \cite{chen2021scene}. Here the recognizer can be replace by any other 2-D attention-based recognizer. In order to make the 1-D attention-based recognizers (\textit{e.g.}, ASTER \cite{shi2018aster}, MORAN \cite{luo2019moran}, and SEED \cite{qiao2020seed}) benefit from the proposed PRAB, we can simply replace the decoder of them with a 2-D attentional decoder.

\subsection{Pluggable Radical-Aware Branch}
By a position-wise multiplication between the feature maps $\mathbf{F}_\text{text}$ extracted by the shared feature extractor and the attention masks $\mathbf{M}_\text{att}$  from the recognition branch, we can obtain the feature maps of each character $\mathbf{F}_\text{char}$ whose dimension is the same as that of $\mathbf{F}_\text{text}$, $\textit{i.e.}$, $\frac{H}{4} \times \frac{W}{4} \times C$. However, more useless features are contained in $\mathbf{F}_\text{char}$. In order to retain the related features of each character, we employ a $1 \times 1$ convolution layer to compress the features $\mathbf{F}_\text{char} \in \mathbb{R}^{\frac{H}{4} \times \frac{W}{4} \times C}$ into $\mathbf{F}^{\prime}_\text{char}$ with the size of $\frac{H}{4} \times \frac{H}{4} \times C$. Subsequently, the compressed features of each character are decoded into the corresponding radical sequence. Benefiting from the proposed PRAB, the shared feature extractor will capture more fine-grained radical-level features for better character recognition. The PRAB, which is only used at the training stage, will not introduce additional inference time.

\subsection{Experimental Results}
As shown in Table \ref{tab:with-prab}, equipped with the proposed PRAB, the performance of attention-based methods is further improved, resulting from that PRAB introduces the more fine-grained supervision (\textit{i.e.}, radicals) to classify characters. For example, with the help of PRAB, the average accuracy improvement on MORAN, SAR, and TransOCR is 1.83\%, 3.35\%, and 2.33\% respectively.

\begin{table*}[ht]
\small
\caption{The experimental results of baseline models that are equipped with the proposed PRAB.}
\centering
\scalebox{1.0}{\begin{tabular}{p{2.5cm}<{\centering} p{1.7cm}<{\centering} p{1.7cm}<{\centering} p{1.7cm}<{\centering} p{1.7cm}<{\centering} p{1cm}<{\centering} p{1cm}<{\centering}}
\toprule
\multirow{2}*{Method} & \multicolumn{4}{c}{Dataset} \\
\cmidrule{2-5}
~  & Scene & Web & Document & Handwriting \\
\midrule
CRNN \cite{shi2016end} & 54.94 / 0.742 & 56.21 / 0.745 & 97.41 / 0.995 & 48.04 / 0.843 \\

\midrule
ASTER \cite{shi2018aster} & 59.37 / 0.801 & 57.83 / 0.782 & 97.59 / 0.995 & 45.90 / 0.819 \\
ASTER + \textit{PRAB} & \textit{60.66} / \textit{0.808} & \textit{58.50} / \textit{0.789} & \textit{\textbf{98.14}} / \textit{\textbf{0.997}} & \textit{48.51} / \textit{0.833} \\

\midrule
MORAN \cite{luo2019moran} & 54.68 / 0.710 & 49.64 / 0.679 & 91.66 / 0.984 & 30.24 / 0.651 \\
MORAN + \textit{PRAB} & \textit{54.43} / \textit{0.708} & \textit{51.85} / \textit{0.701} & \textit{91.87} / \textit{0.984} & \textit{35.37} / \textit{0.712} \\

\midrule
SAR \cite{li2019show} & 53.80 / 0.738 & 50.49 / 0.705 & 96.23 / 0.993 & 30.95 / 0.732 \\
SAR + \textit{PRAB} & \textit{55.10} / \textit{0.740} & \textit{53.86} / \textit{0.722} & \textit{97.10} / \textit{0.994} & \textit{38.81} / \textit{0.782} \\

\midrule
SEED \cite{qiao2020seed} & 45.37 / 0.708 & 31.35 / 0.571 & 96.08 / 0.992 & 21.10 / 0.555  \\
SEED + \textit{PRAB} & \textit{49.89} / \textit{0.739} & \textit{31.57} / \textit{0.572} & \textit{97.77} / \textit{0.996} & \textit{21.38} / \textit{0.556} \\

\midrule
TransOCR \cite{chen2021scene} & 67.81 / 0.817 & 62.74 / 0.782 & 97.86 / 0.996 & 51.67 / 0.835  \\
TransOCR + \textit{PRAB} & \textit{\textbf{71.02}} / \textit{\textbf{0.843}} & \textit{\textbf{63.82}} / \textit{\textbf{0.794}} & \textit{98.02} / \textit{\textbf{0.997}} & \textit{\textbf{56.54}} / \textit{\textbf{0.869}} \\
\bottomrule
\end{tabular}
}
\label{tab:with-prab}
\end{table*}

\section{Pretrained with Synthetic Datasets}
Most of existing scene text recognizers tend to be pretrained with a large scale of synthetic dataset. Thus, for Chinese text recognition, we also generate a synthetic training dataset, which contains 20M samples, by the text generator \footnote{https://github.com/Belval/TextRecognitionDataGenerator} with the corpus of Sougou news \footnote{https://github.com/lijqhs/text-classification-cn}. We first use the synthetic dataset to pretrain baseline models, and then fine tune baseline models on the four collected datasets. The experimental results (shown in Table \ref{tab:syn}) demonstrate that pretraining with the generated synthetic dataset can indeed improve the performance of baseline models in most cases. However, these baseline models designed for English text recognition still do not achieve the expected performance.

\begin{table*}[ht]
\small
\caption{The experimental results of baseline models that are pretrained with the synthetic dataset.}
\centering
\scalebox{1.0}{\begin{tabular}{p{2.5cm}<{\centering} p{1.7cm}<{\centering} p{1.7cm}<{\centering} p{1.7cm}<{\centering} p{1.7cm}<{\centering} p{1cm}<{\centering} p{1cm}<{\centering}}
\toprule
\multirow{2}*{Method} & \multicolumn{4}{c}{Dataset} \\
\cmidrule{2-5}
~  & Scene & Web & Document & Handwriting \\
\midrule
ASTER \cite{shi2018aster} & 59.37 / \textit{0.801} & 57.83 / 0.782 & 97.59 / 0.995 & 45.90 / 0.819 \\
ASTER + \textit{Synth} & \textit{60.25} / 0.796 & \textit{\textbf{63.35}} / \textit{\textbf{0.804}} & \textit{\textbf{98.77}} / \textit{\textbf{0.998}} & \textit{54.39} / \textit{0.866} \\

\midrule
SEED \cite{qiao2020seed} & 45.37 / 0.708 & 31.35 / 0.571 & 96.08 / 0.992 & 21.10 / 0.555  \\
SEED + \textit{Synth} & \textit{58.47} / \textit{0.787} & \textit{62.76} / \textit{0.801} & \textit{98.44} / \textit{0.997} & \textit{\textbf{55.19}} / \textit{\textbf{0.869}} \\

\midrule
SAR \cite{li2019show} & 53.80 / 0.738 & 50.49 / 0.705 & 96.23 / 0.993 & 30.95 / 0.732 \\
SAR + \textit{Synth} & \textit{57.96} / \textit{0.761} & \textit{57.67} / \textit{0.745} & \textit{97.56} / \textit{0.996} & \textit{39.24} / \textit{0.798} \\

\midrule
MASTER \cite{li2019show} & \textit{62.14} / \textit{0.763} & 53.42 / 0.704 & 82.69 / 0.957 & 18.52 / 0.504 \\
MASTER + \textit{Synth} & 60.96 / 0.748 & \textit{58.72} / \textit{0.739} & \textit{92.85} / \textit{0.984} & \textit{30.36} / \textit{0.678} \\

\midrule
ABINet \cite{li2019show} & 60.88 / 0.775 & \textit{51.07} / \textit{0.704} & 91.67 / 0.987 & 13.83 / 0.514 \\
ABINet + \textit{Synth} & \textit{62.45} / \textit{0.786} & 49.61 / 0.699 & \textit{92.86} / \textit{0.989} & \textit{15.53} / \textit{0.529} \\

\midrule
TransOCR \cite{chen2021scene} & 67.81 / 0.817 & \textit{62.74} / \textit{0.782} & 97.86 / 0.996 & 51.67 / 0.835  \\
TransOCR + \textit{Synth} & \textit{\textbf{68.54}} / \textit{\textbf{0.824}} & 62.54 / 0.781 & \textit{97.90} / \textit{0.997} & \textit{53.48} / \textit{0.848} \\

\bottomrule
\end{tabular}
}
\label{tab:syn}
\end{table*}

\section{Visualization of Failure Cases.}
\begin{figure*}[ht]
    \centering
    \includegraphics[width=1\textwidth]{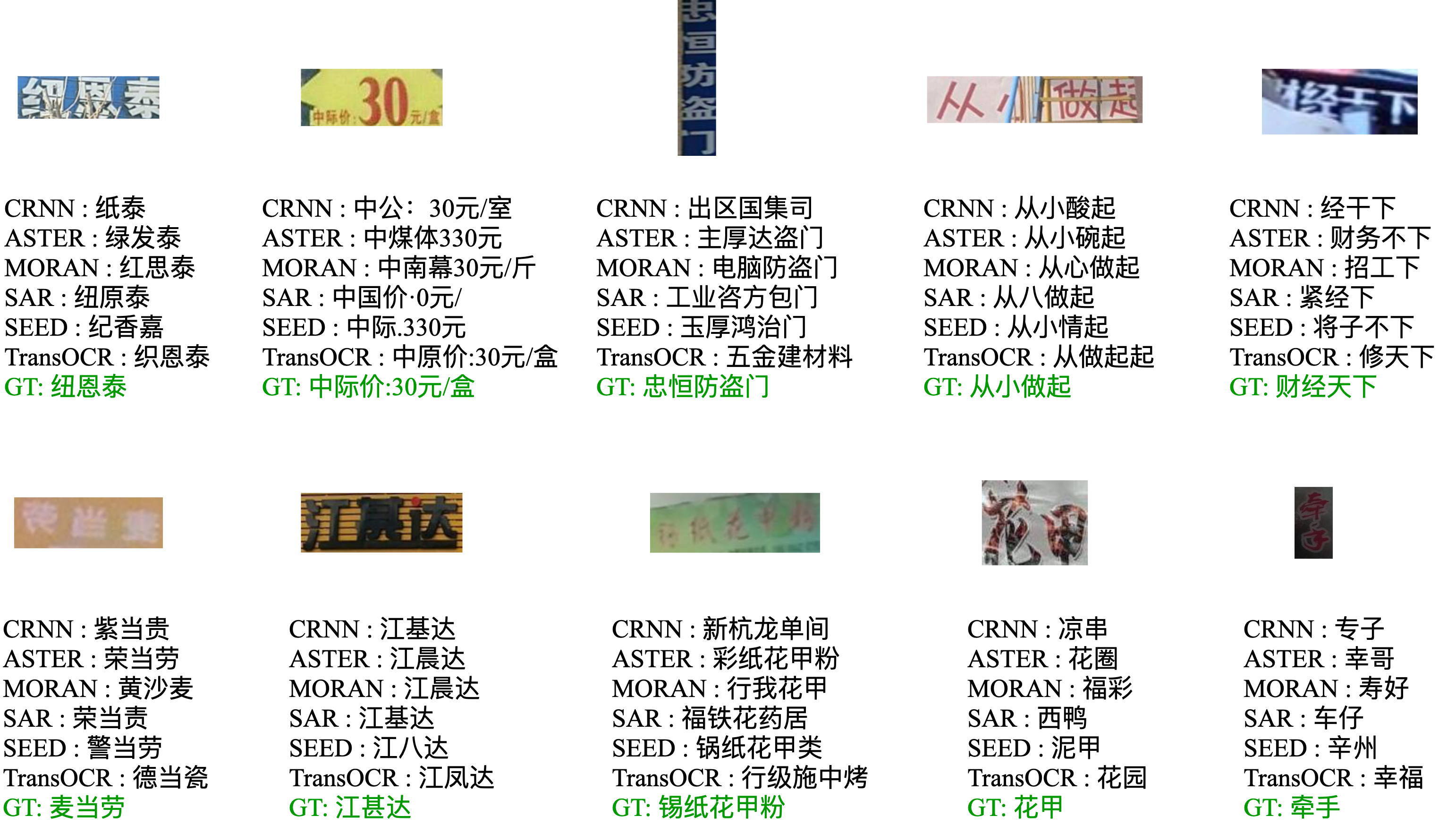}
    \caption{Failure cases in the scene dataset.}
    \label{fig:scene failure}
\end{figure*}

\begin{figure*}[ht]
    \centering
    \includegraphics[width=1\textwidth]{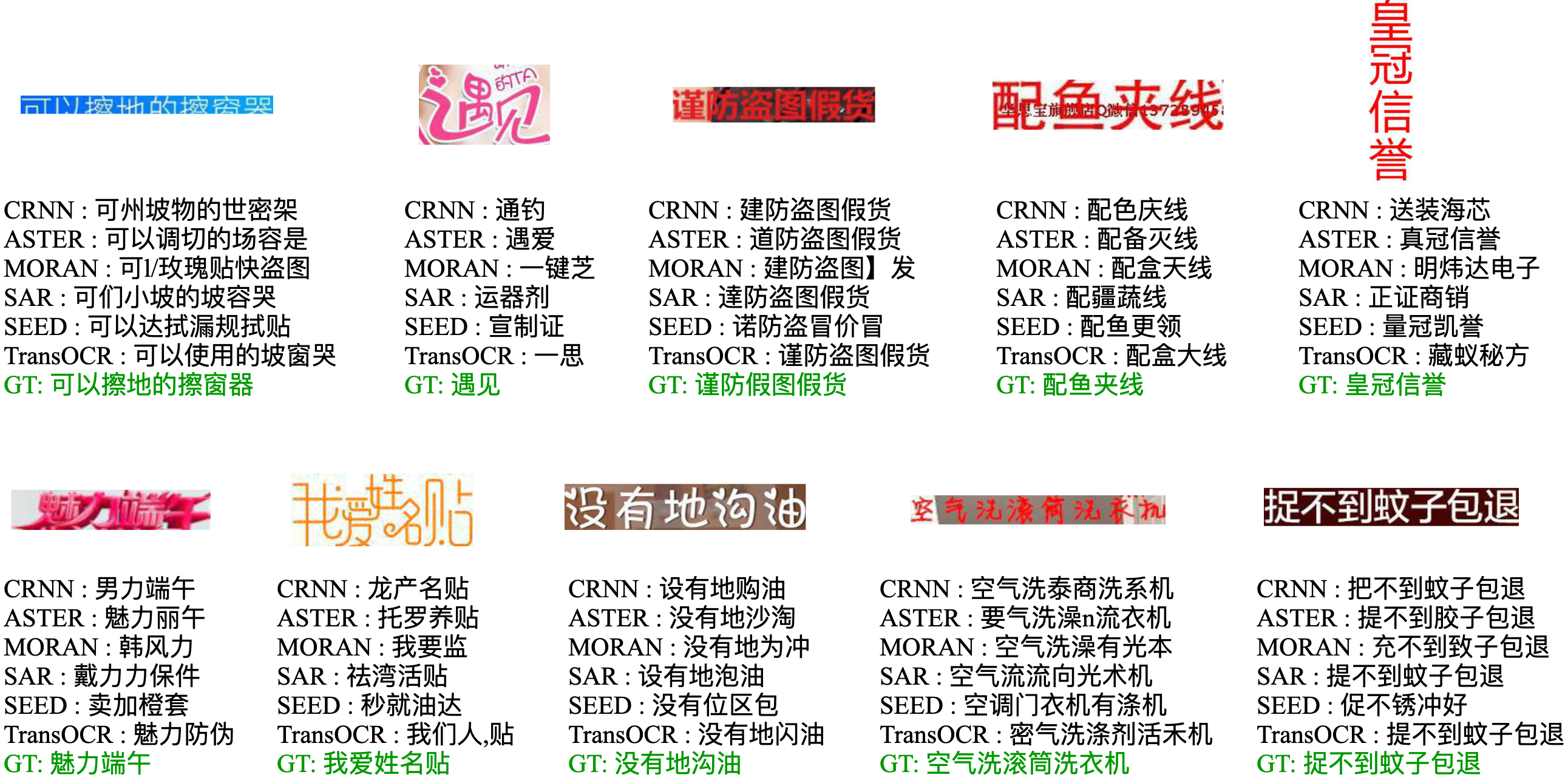}
    \caption{Failure cases in the web dataset.}
    \label{fig:web failure}
\end{figure*}

\begin{figure*}[ht]
    \centering
    \includegraphics[width=1\textwidth]{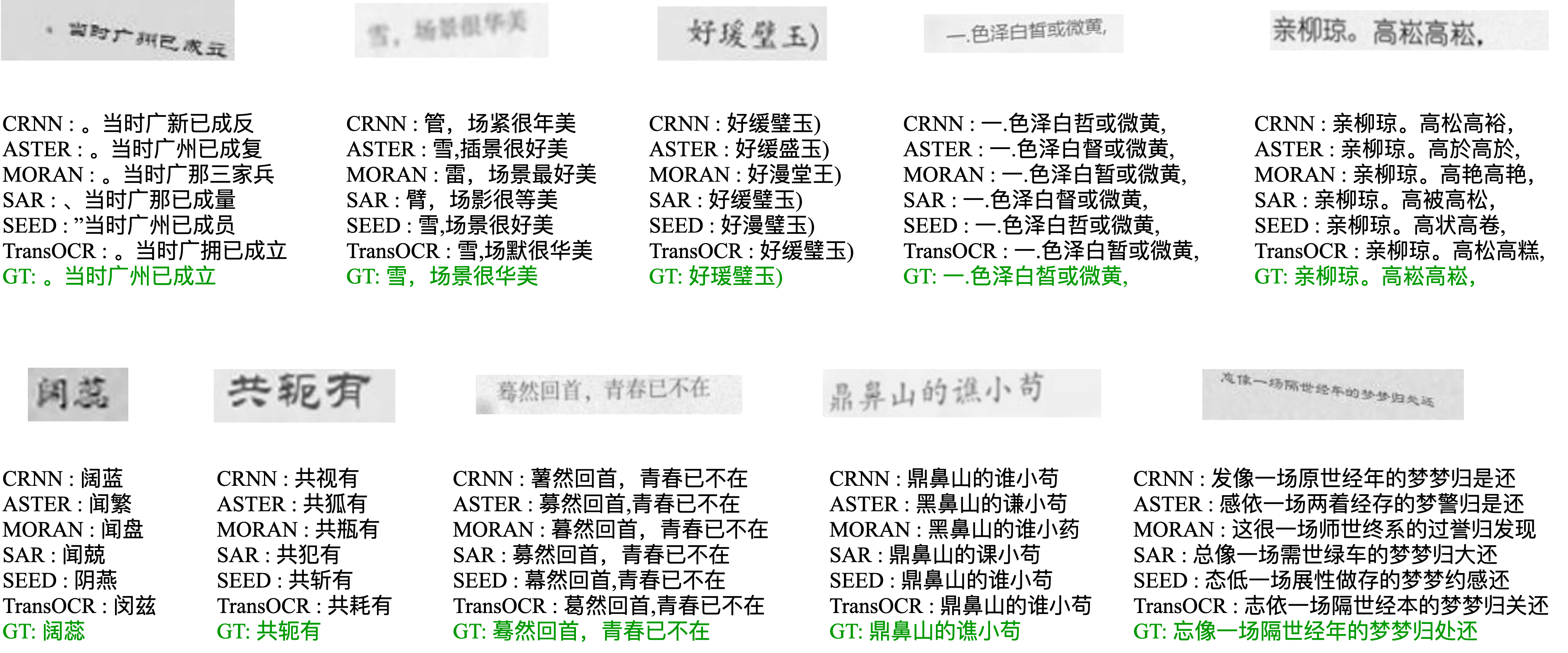}
    \caption{Failure cases in the document dataset.}
    \label{fig:document failure}
\end{figure*}

\begin{figure*}[ht]
    \centering
    \includegraphics[width=1\textwidth]{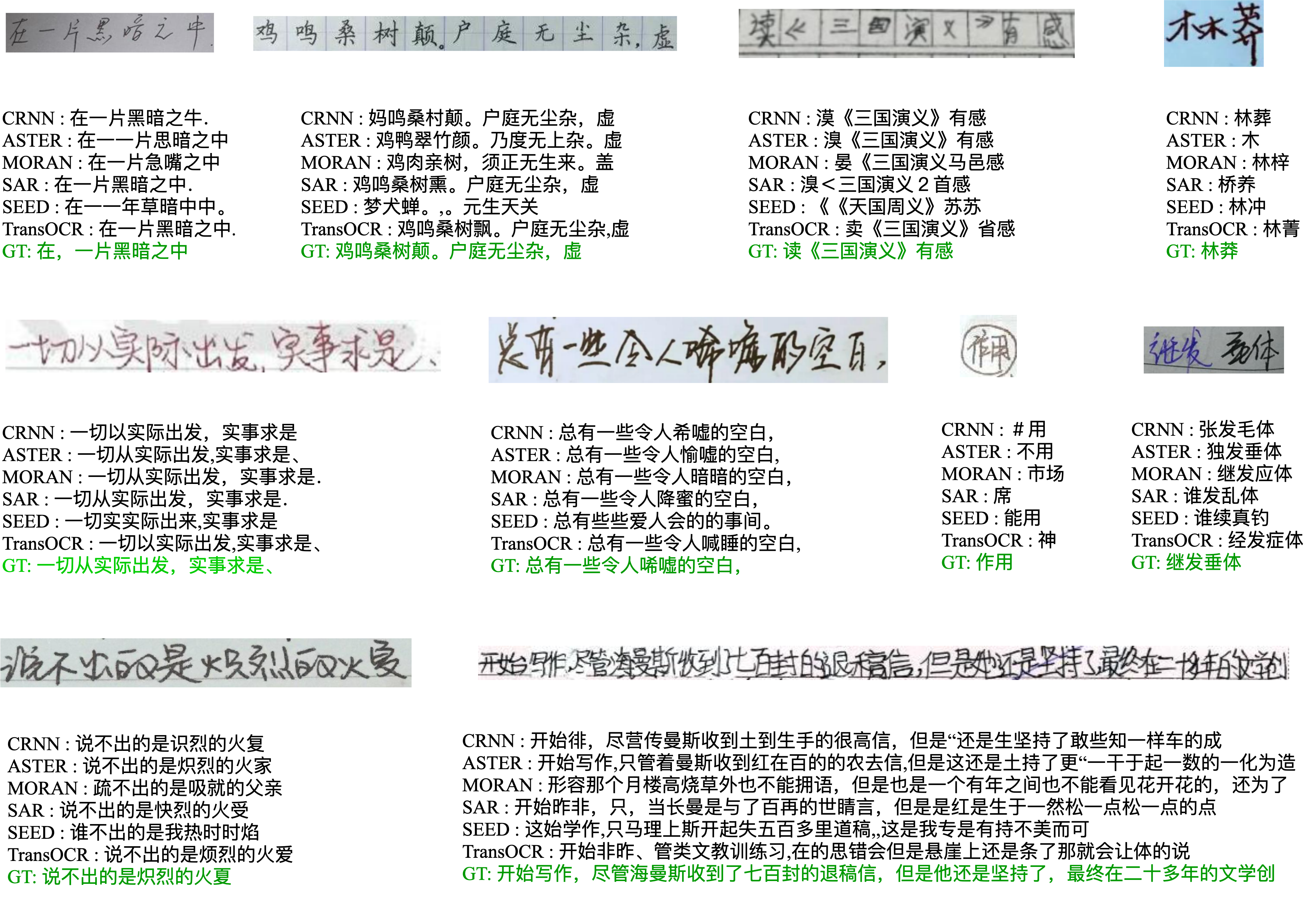}
    \caption{Failure cases in the handwriting dataset.}
    \label{fig:handwriting failure}
\end{figure*}

We have visualized some failure cases in Figure \ref{fig:scene failure}, Figure \ref{fig:web failure}, Figure \ref{fig:document failure}, and Figure \ref{fig:handwriting failure} for the scene, web, document, handwriting datasets, respectively. In particular, we pick ten samples from each dataset that are wrongly predicted by all the baselines. As demonstrated in Figure \ref{fig:scene failure}, we notice that the occluded texts (\textit{e.g.}, \ch{``纽恩泰''}, \ch{``财经天下''}, \ch{``冰淇淋鸡蛋仔''}, and \ch{``从小做起''}) indeed bring difficulties to the recognizers as the foregrounds may be mistakenly perceived as parts of the texts. Moreover, there are some extremely hard cases like mirrored texts (\textit{e.g.}, \ch{``麦当劳''}), which are illegible even for human eyes. For the failure cases in the web dataset, we notice that the baselines are hard to tackle text images with artistic fonts (\textit{e.g.}, \ch{``遇见''}, \ch{``魅力端午''},  \ch{``没有地沟油''}, and \ch{``我爱姓名贴''}). As shown in Figure \ref{fig:document failure}, although the document dataset is of the highest recognizability as it is synthesized by text render, some samples still pose difficulties to all baselines. We notice the baselines fail to tackle some rarely-used characters (\textit{e.g.}, \ch{``瑗''}, \ch{``菘''}, and \ch{``轭''}), thus mistakenly regarding them as other similar characters. For the handwriting dataset, we notice that the join-up or missing strokes may confuse the baselines (\textit{e.g.}, \ch{`垂体''} and \ch{``火夏''}). Additionally, some images with long texts may make it hard for baselines to capture the sequential patterns.


\clearpage

\end{document}